\documentclass[12pt]{amsart}
\usepackage{amsbsy,amssymb,amsmath,amsthm,amscd,amsfonts,latexsym,amstext,delarray,
amsmath,graphicx} 
\usepackage[margin=.8in]{geometry}
\usepackage{color}

\numberwithin{equation}{section}

\def\C{{\mathbb C}}

\def\P{{\mathbb P}}

\def\cI{{\mathcal I}}

\title{Syntactic Phylogenetic Trees}
\author[K.Shu, S.Aziz, V.L.Huynh, D.Warrick, M.Marcolli]{Kevin Shu, Sharjeel Aziz, Vy-Luan Huynh, David Warrick, Matilde Marcolli}
\address{Division of Physics, Mathematics, and Astronomy \\ California Institute of Technology \\
1200 E. California Blvd, Pasadena, CA 91125, USA}

\email{kshu@caltech.edu}
\email{saziz@caltech.edu}
\email{vhuynh@caltech.edu}
\email{warrick.david58@gmail.com}
\email{matilde@caltech.edu}
\date{}

\begin{document}
\maketitle

\begin{abstract}
In light of recent controversies surrounding the use of computational methods for the
reconstruction of phylogenetic trees of language families (especially the Indo-European
family), a possible approach based on syntactic information, complementing other linguistic methods,
appeared as a promising possibility, largely developed in recent years in Longobardi's Parametric 
Comparison Method. In this paper we identify several serious problems that arise in the
use of syntactic data from the SSWL database for the purpose of computational
phylogenetic reconstruction. We show that the most naive approach fails to produce 
reliable linguistic phylogenetic trees. We identify some of the sources of the
observed problems and we discuss how they may be, at least partly, corrected by using additional
information, such as prior subdivision into language families and subfamilies, and a better
use of the information about ancient languages. We also describe how the use of phylogenetic 
algebraic geometry can help in estimating to what extent the probability distribution at the
leaves of the phylogenetic tree obtained from the SSWL data can be considered reliable,
by testing it on phylogenetic trees established by other forms of linguistic analysis. In simple
examples, we find that, after restricting to smaller language subfamilies and considering only
those SSWL parameters that are fully mapped for the whole subfamily, the SSWL data match
extremely well reliable phylogenetic trees, according to the evaluation of phylogenetic invariants.
This is a promising sign for the use of SSWL data for linguistic phylogenetics. 
We also argue how dependencies and nontrivial geometry/topology in the space of
syntactic parameters would have to be taken into consideration in phylogenetic reconstructions
based on syntactic data. A more detailed analysis of syntactic phylogenetic trees and their 
algebro-geometric invariants will appear elsewhere.
\end{abstract}

\section{Introduction}

This paper is based on a talk given by the last author at the workshop ``Phylogenetic Models:  Linguistics, Computation, and Biology" organized by Robert Berwick at the CSAIL department of MIT in May 2016.

\smallskip

The reconstruction of phylogenetic trees of language families is a crucial 
problem in the field of Historical Linguistics. The construction of an
accurate family tree for the Indo-European languages  
accompanied and originally motivated the development of Historical Linguistics,
and has been a focus of attention for linguists for the span of two centuries. In recent
years, Historical Linguistics has seen a new influx of mathematical and computational methods, 
originally developed in the context of mathematical biology to deal with species 
phylogenetic trees, see for instance 
\cite{BHGK}, \cite{DelCri}, \cite{Forster}, \cite{PeSe}, \cite{Nak}, \cite{Warnow}.
A considerable amount of controversy arose
recently in relation to the accuracy and effectiveness of these methods and
the related problem of phylogenetic inference. In particular, claims regarding
the phylogenetic tree of the Indo-European languages made in \cite{Bouck} 
were variously criticized by historical linguists, see the detailed discussion in \cite{PerLe}.

\smallskip

Most of the literature dealing with computational phylogenetic trees in the context
of Linguistics focused on the use of lexical data, in the form of Swadesh lists of
words, and the encoding as binary data of the counting of cognate words, see for
instance the articles in \cite{Forster}. Other reconstructions used phonetic data
and sound change, as in \cite{BHGK}, or a combination of several types of
linguistic data (referred to as ``characters"), including phonetic, lexical, and
morphological properties, as in \cite{Barba}, \cite{Warnow}. A different approach
to linguistic phylogenetic reconstruction, based on syntactic parameters, was
developed recently in \cite{Longo}, \cite{Longo2}, \cite{Longo3}, \cite{LongGua},
\cite{LongGua2}. This method is known as Parametric Comparison Method (PCM).  
A coding theory perspective on the PCM was given in \cite{Mar}.

\smallskip

The notion of syntactic parameters arises in Generative Linguistics, within the
Principles and Parameters model developed by Chomsky in \cite{Chomsky},
\cite{ChoLa}. A more expository account of syntactic parameters is given
in \cite{Baker}. Syntactic parameters are conceived as binary variables that
express syntactic features of natural languages. The notion of syntactic
parameters has undergone changes, reflecting changes in the modeling 
of generative grammar: for a recent overview of the parametric modeling
of morphosyntactic features see \cite{Rizzi}. 
A main open problems in the parameteric approach for comparative generative 
grammar is  understanding the space of syntactic parameters, identifying
dependence relations between parameters and possibly identifying a 
fundamental set of such variables that would represent a good system of
coordinates for the space of languages. Recently, the use of mathematical
methods for the study of the space of syntactic parameters of world languages
was proposed in \cite{PBZMYKM}, \cite{PGGCLDM}, \cite{STM}. 

\smallskip

At present, the only available
extensive database of binary parameters describing syntactic features is
the SSWL database \cite{SSWL}, which collects data of 115 parameters
over 253 world languages. It is debatable whether the binary variables
collected in SSWL represent fundamental syntactic parameters:  
surface orders, for instance, are often confounded with the deep underlying parameter values.
Moreover, SSWL does not record any dependence relations between 
parameters. Different data of syntactic parameters have been used 
in \cite{LongGua}, \cite{LongGua2}, with dependence relations taken
into account, and more data are being collected by these authors and 
will hopefully be available soon. 
For the purpose of this paper, we will use the terminology ``syntatic parameters" 
loosely for any collection of binary variables describing syntactic features of 
natural languages. We work with the SSWL data, simply because it is presently 
the most extensive database available of syntactic structures. 

\smallskip 

In Section \ref{PHYLIPsec} of this paper we show that just using the Hamming distance between vectors of
binary variables extracted from the SSWL data and the Neighborhood-Joining Method
for phylogenetic inference gives very poor results as far as linguistic phylogenetic
trees are concerned. We identify several different sources of problems, some
inherent to the SSWL data, some to the inference methodology, and some more
generally related to the use of syntactic parameters for phylogenetic linguistics. 

\smallskip

In the Section \ref{AGsec} we review the method of Phylogenetic Algebraic Geometry
of \cite{PaSturm} and the main results of \cite{AllRho} and \cite{SturmSull} on 
phylogenetic ideals and phylogenetic invariants that we need for applications to
the analysis of syntactic phylogenetic trees. In Section \ref{AGtreesec} 
we show how one
can use techniques from Phylogenetic Algebraic Geometry to test the reliability of
syntactic parameter data for phylogenetic linguistics, by using known phylogenetic 
trees that are considered reliable, and to test the reliability of candidate 
phylogenetic trees assuming a certain degree of reliability of the syntactic data. 

\smallskip

In Section \ref{GeomSec} we argue that dependencies between the syntactic
variables recorded in the SSWL database should be taken into consideration
in order to improve the reliability of these data for phylogenetic reconstruction.
In particular, the presence of geometry/topology in this set of data and the
presence of different degrees of recoverability of some of the SSWL syntactic 
variables in Kanerva network tests indicate that an appropriated weighted use
of the data that accounts for these phenomena may improve the results. 

\smallskip

\subsection*{Acknowledgment} The first author is supported by a Summer
Undergraduate Research Fellowship at Caltech. 
Part of this work was performed as part of the activities of the last author's 
Mathematical and Computational Linguistics lab and CS101/Ma191 class at Caltech. The last author 
is partially supported by NSF grants DMS-1201512 and PHY-1205440. 

\bigskip

\section{PHYLIP analysis of SSWL}\label{PHYLIPsec}

We discuss here the problems that occurs in a naive analysis of the SSWL database
using the phylogenetic tree algorithm PHYLIP. We identify the main types of
errors that occur and the possible sources of the problems. We will discuss in
\S \ref{AGsec} how one can eliminate some of the problems and obtain more
accurate phylogenetic trees from SSWL data, using different methods.

\subsection{Data and Code}

We acquired the syntactic language data from the SSWL database with two different methods, one consisting of downloading the data as a {\em .csv} file directly, with the results separated in the format {\em ``language$|$property$|$value''}, and one achieved by scraping the data into a {\em .json} file, formatted as a list of lists of binary variables, in the format ``{\em `language' : $\{$`parameters' : `values'$\}$}". 
This was done with a python script 
{\tt \verb!data_obtainer.py!} which went through all of SSWL and dumped the data as desired.

\smallskip

The SSWL data, stored in a more convenient {\tt .json} file format produced by the first author, are
available as the file {\tt full\_langs.json} which can be downloaded at the URL address \newline
 {\tt http://www.its.caltech.edu/$\sim$matilde/PhylogeneticSSWL2}.

\smallskip

We created, for each language in the database, a vector of binary variables representing the syntactic traits
of that language as recorded in the SSWL database, with value $1$ indicated that the language 
possessed the respective trait, and value $0$ indicating that the language does not possess the trait. 

\smallskip

One of the main sources of problems regarding the use of SSWL data arises already at this stage: not all
languages in the database have all the same parameters mapped. The lack of information about 
a certain number of parameters for certain languages alters the counting of the Hamming distances,
as it requires a choice of normalization of the string length, with additional entries added representing
lack of information. This clearly generates problems, as this inconsistency generates mistakes in
the counting of Hamming distances and in the tree reconstruction. In \S \ref{TreeSec} we will illustrate 
specific examples where this problem occurs. 

\smallskip

The Hamming distance algorithm {\tt HF.py} takes two equal-length binary sequences, throwing an 
error if this length requirement is violated, and returns the sum of all bitwise XORs between them, 
or the total number of differences. In this way, we construct with {\tt \verb! distance_matrix_checker.py!}
the Hamming distance matrix  $M_{ab}=d_H(\ell_a,\ell_b)$, whose entries are the Hamming 
distances between the vectors of binary syntactic parameters of languages $\ell_a$ and $\ell_b$. 

\smallskip

For example, Germanic languages on average have normalized Hamming distance in the range 
$0.3$-$0.4$. Old Saxon and Old English have a Hamming distance of $0.17$ from German, while
Swiss German has distance $0.09$. Modern English has below average differences at $0.27$. 
While these distances may appear reasonable, 
one can detect easily another major source of problems in the use of SSWL data for 
phylogenetic reconstruction.
Many languages belonging to very different families have small Hamming distance: for example, the
Indo-European Hindi (60$\%$ mapped in SSWL) and the Sino-Tibetan Mandarin 
(87$\%$ mapped in SSWL) receive a normalized distance of $0.12$. This is certainly in large
part due to the different level of accuracy with which the two languages are mapped in the same 
database. However, one can also observe syntactic similarities between languages belonging to
different families, which are not due to poor recording of the respective data, but are a genuine
consequence of the syntactic properties being described. 

\smallskip

This $253\times 253$ matrix of Hamming distances was then given as input to the PHYLIP 
package\footnote{{\tt http://evolution.genetics.washington.edu/phylip/software.html}} 
for phylogenetic tree reconstruction, which is widely used in Mathematical Biology.
Given the Hamming distance matrix $M_{ab}=d_H(\ell_a,\ell_b)$, 
the PHYLIP software provides several options for
tree construction from distance matrix data: additive tree model, ultrametric model,
neighbor joining method, and average linkage clustering (UPGMA). The resulting 
tree produced by PHYLIP, containing all 253 languages in the SSWL database, is contained in 
{\tt outfile}, where the tree in the text file is drawn with dashes and exclamation points.
The information of the output tree and distances is also given in the 
output file {\tt outtree} in Newick format, with parentheses and commas. 
The accompanying file {\tt key.txt} contains the key that indicates the full language 
name that corresponds to each two-letter string in outfile. The output files can be opened in 
any text editor.

\smallskip

The python code and the output files, prepared by the second, third and fourth
authors of this paper, are available at 
{\tt http://www.its.caltech.edu/$\sim$matilde/PhylogeneticSSWL}

\subsection{Main problems in the resulting tree}\label{TreeSec}

A quick inspection of the output file obtained by running PHYLIP
on the SSWL data immediately reveals that there are many problems
with the resulting phylogenetic tree. We will give explicit examples here
that illustrate some of the main type of problems one encounters.
There are many more such examples one can easily find by
inspecting the output tree available in the repository at the URL
indicated above. 

\subsection{Sources of problems}\label{ListSec}
  
An important problem in computational phylogenetic reconstruction is 
how to validate statistically the model. There are well known problem inherent
in using the Hamming distance as a source for phylogenetic trees.
Estimating tree branch lengths is a hard problem. Distance matrices can be
non-additive due to error, and it is typically difficult to distinguish
distances that deviate from additivity due to change from  
deviations due to error. This problem is significant even in the context 
of Biology, where the use of DNA data is more reliable than the use of
vectors of binary variables coming from linguistic properties. For a
discussion of some of these issues in Biology see \cite{DeDe}. For
a comparison of phylogenetic methods (not including syntactic
parameters) in Linguistics, see \cite{Barba}. 

\smallskip

As we discuss with individual specific examples in the subsections 
that follow, there are several different source of problems that 
combine to create different kinds of errors in the resulting
phylogenetic tree. The main problems are the following:
\begin{enumerate}
\item inherent problems in the computational method based on Hamming distances, as
discussed above;
\item problems with non-uniform coverage of syntactic data across different languages 
and language families in the SSWL database;
\item the nature of the syntactic variables recorded in the SSWL database (for instance
with respect to surface versus deep structure) and the presence of relations between
these variables;
\item the existence of languages belonging to unrelated linguistic families that can
be similar at the level of syntactic structures.
\end{enumerate}

Clearly, some of these problems are of linguistic nature, like the last one listed,
while others are of computational nature, like the first one, while others depend
on the nature and accuracy of the SSWL data. It is difficult to disentangle the
effects of each individual problem on the output tree, but the examples listed
below illustrate cases where one can identify one of the problems listed here as
the most likely origin of the mistakes one sees in the resulting phylogenetic tree.

\subsubsection{Misplacement of languages within the correct subfamily tree}\label{wrong1sec}
This type of problem occurs when a group of languages are correctly identified
as belonging to the same subfamily of a given historical-linguistic family, but
the internal structure of the subfamily tree appears inconsistent with the
structure generally agreed upon based on other linguistic data. 

\smallskip

\begin{center}
\begin{figure}
\includegraphics[scale=0.85]{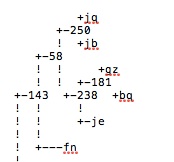}
\caption{PHYLIP output subtree of Latin languages: jq=Spanish, jb=Catalan, gz=Sicilian, bq=Italian,
je=Portuguese, fn=French. \label{LatinFig}}
\end{figure}
\end{center}

In the naive PHYLIP
analysis of the SSWL database we see an example of this kind by considering the
subtree of the Latin languages within the Indo-European family. The shape of this
subtree, as it appears in in the output file, is of the form illustrated in Figure~\ref{LatinFig}.
We see here that, although these languages are correctly grouped together as
belonging to the same subfamily, the relative position within the subtree does not
agree with what historical linguistic methods have established. Indeed, one can
easily see, for instance, that the position of Portuguese in the subtree is incorrectly
placed closer to Italian and Sicilian, than to Spanish and Catalan. This example
is interesting because the error does not appear to be due to the poor mapping
of parameters for these languages: Italian  and Sicilian are $100\%$ mapped in SSWL
and Spanish, Catalan, and Portuguese are $84\%$ mapped. So these are among
some of the best recorded languages in the database, and still their respective
position in the phylogenetic tree does not agree with reliable reconstructions from
Historical Linguistics. It is interesting to compare the reconstruction obtained in
this way, with the one obtained, on a different set of syntactic data, by Longobardi's
Parametric Comparison in \cite{LongGua}, which has Italian and French as a pair of two
nearby branches, and Spanish and Portuguese as another pair of nearby branches. 
This example appears to outline an issue arising from the way syntactic variables
are classified in the SSWL (as opposed to the different list of syntactic parameters
used in \cite{LongGua}). We discuss in \S \ref{GeomSec} below some of the
problems of dependencies between the SSWL syntactic variables that may be
at the sources of this kind of problem.

\subsubsection{Placement of languages in the wrong subfamily tree}\label{wrong2sec}

\begin{center}
\begin{figure}
\includegraphics[scale=0.85]{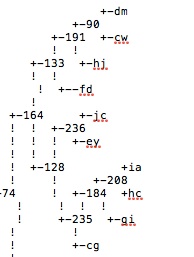}
\caption{PHYLIP output subtree of Germanic languages: 
dm=Norwegian, cw=Faroese, hj=Italian Ancient Neapolitan, fd=Icelandic, jc=Afrikaans, 
ey=West Flemish, ia=Dutch, hc=German, gi=Swedish, cg=English. \label{GermFig}}
\end{figure}
\end{center}

Another type of mistake one finds in the naive phylogenetic tree reconstruction
from SSWL syntactic data is illustrated by the Germanic languages in Figure~\ref{GermFig}.
In this case, we find that most of the languages in this subtree are correctly grouped together 
as Germanic, but a language that clearly belongs to a different subfamily is also placed in
the same group. It is very puzzling why Ancient Neapolitan ends up incorporated in
the tree of Germanic languages rather then near Italian and the other dialects of Italian
in the subtree of Latin languages of Figure~\ref{LatinFig}. Linguistically, one could perhaps
argue that Ancient Neapolitan did in fact have several Germanic influences due to the
Ostrogoths, but it is more reasonable to expect such influences to appear at the
lexical rather than syntactic level. Moreover, the specific placement within the Germanic
tree near Faroese, Norwegian and Icelandic, does not necessarily reflect this hypothesis.
In terms of the accuracy with which these languages are recorded in the SSWL database,
Ancient Neapolitan is $83\%$ mapped, while its nearest neighbor on this PHYLIP output tree
have Norwegian, which is also mapped with a similar accuracy of $84\%$, and 
Faroese and Icelandic with a lower accuracy of $69\%$. It is possible that this example
already reflects a problem with the different accuracy of mapping of different languages
in the SSWL database, or it may be a problem with the algorithmic reconstruction method
itself. There are several similar instances in the output tree, which point to a problem that
is systematic, hence likely generated by the method of phylogenetic reconstruction adopted
in this naive analysis. 

\subsubsection{Proximity of languages from unrelated families}\label{proxSec}

\begin{center}
\begin{figure}
\includegraphics[scale=0.85]{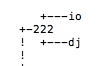}
\caption{Misplaced proximity: io=K'iche', dj=Georgian. \label{MayaFig}}
\end{figure}
\end{center}

Another type of problem that occurs frequently in the output tree of this
naive analysis is the case of completely unrelated languages (from 
completely different language families) that are placed in adjacent
positions in the tree. We see an example in Figure~\ref{MayaFig},
where the Mayan K'iche' language and Georgian (Kartvelian family)
are placed next to each other in the tree. 
Both K'iche' and Georgian are $69\%$ mapped in the SSWL database.
Although this is not as accurate a mapping as some of the languages 
we discussed in the previous examples, it is nonetheless the same
level of precision available, for instance, for some of the Germanic
languages in the previous example, which were at least placed
correctly in the Germanic subtree. Thus, the type of problem we see
in this example is not entirely due to poor mapping of the languages
involved. It must be also an effect of other factors like the 
computational reconstruction method used, as in the previous class
of examples. However, there can also be some purely linguistic
factors involved. Namely, there are known cases of languages
belonging to unrelated historical linguistic families that may appear
close at the syntactic level. This type of phenomenon may be
responsible for at least part of the cases where one finds unrelated
languages placed in close proximity in the output tree. This is an
indication that one should not rely on syntactic data alone, without
accompanying them with other linguistic data, that can provide,
for example, a prior subdivision of languages into language families.
Using the same method of phylogenetic tree reconstruction on
data already grouped into linguistic families, with individual family
trees separately constructed, improves the accuracy of the
resulting trees. Other combinations of syntactic and lexical/morphological
data can be used to improve accuracy.

\subsubsection{The position of ancient languages in the tree}\label{wrong4sec}

\begin{center}
\begin{figure}
\includegraphics[scale=0.75]{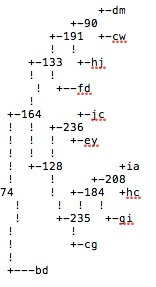} 
\caption{The position of Old English with respect to the Germanic languages:
bd= Old English. \label{OldEnglFig}}
\end{figure}
\end{center}

\begin{center}
\begin{figure}
\includegraphics[scale=0.75]{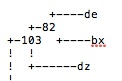}
\caption{Proximity of Ancient Greek and Latin:  
de=Latin, bx=Ancient Greek,  dz=Medieval Greek. \label{AGreekFig}}
\end{figure}
\end{center}

Finally, there is an additional problem one encounters in the naive
phylogenetic reconstruction based on the SSWL data, namely
the position of the ancient languages in the tree. Clearly, the
algorithm assumes that all the data correspond to leaves of the
tree and that the inner nodes are hidden variables, while the
fact that we do have knowledge of some of the ancient languages
and that several are recorded in the SSWL database means that
some of the inner nodes should in fact carry some of the data.
This problem can be resolved if the inner languages would be
placed as a single leaf attached to the corresponding inner node.
By inspecting the resulting output tree we see that sometimes this
is the case, and the inner node to which the corresponding ancient 
language is attached reasonably with respect to the modern
languages that derived from it. One such example is the position
of Old English with respect to the tree of the Germanic languages
in Figure~\ref{OldEnglFig}. However, in other cases, ancient
languages are correctly placed in proximity of each other, but
in the wrong position, in the tree, with respect to the resulting 
modern languages. This is the case with Ancient Greek and
Latin (see Figure~\ref{AGreekFig}). In this case, the algorithm
correctly captures the close syntactic proximity between 
Ancient Greek and Latin, but it does not place these two
languages correctly with respect to either the tree of Latin
languages nor the modern part of the Hellenic branch. This
problem can be improved by first subdividing the data into
language families and smaller subfamilies and then perform
the phylogenetic tree reconstruction on the subfamilies separately,
so that the corresponding ancient language is placed correctly,
and then related the resulting trees by proximity of the ancient
languages. However, this method clearly applies only where
enough other linguistic information is available, in addition
to the syntactic data. It should be noted, moreover, that, while
Ancient Greek is correctly placed in proximity to Latin,
Homeric Greek is entirely misplaced in the PHYLIP tree
reconstruction and does not appear in proximity of the Ancient Greek
of the classical period, even though both Homeric and Ancient
Greek are mapped with the best possible accuracy ($100\%$ mapped) 
in the SSWL database. 

\subsection{The Indo-European tree}

Although the many problems illustrated above render a phylogenetic
reconstruction based solely on SSWL data unreliable, it is still worth
commenting on what one obtains with this method regarding some of the 
controversial early branchings of the Indo-European tree. Again, the same
type of systematic problems illustrated above occur repeatedly when one analyzes
these regions of the output tree.

\medskip

\begin{center}
\begin{figure}
\includegraphics[scale=0.75]{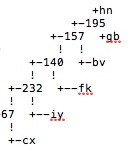}
\caption{Tocharian--Anatolian branching:  
gb=Tocharian A, hn=Tocharian B, 
bv=Hittite, fk=Pashto, iy=Panjabi,
cx=Tuvan (Turkic). \label{TochFig}}
\end{figure}
\end{center}

For example, Tocharian A and B are treated by the PHYLIP reconstruction
as modern languages leaves of the tree and placed in immediate proximity 
of Hittite and in close proximity of some of the modern Indo-Iranic languages,
like Pashto and Punjabi, and a further step away from some Turkic languages
like Tuvan. The proximity of Tocharian and Hittite suggests here a Tocharian-Anatolian
branching.  The placement of the Indo-Iranic languages in proximity of this 
Tocharian-Anatolian branching is likely arising from the fact that the Indo-Iranic
branch of the Indo-European family is very poorly mapped in the SSWL database,
with the ancient languages entirely missing and very few of the modern languages
recorded, hence the reconstructed tree necessarily skips over all these missing
data. The complete absence of Sanskrit from the current version of the SSWL database
(the entry in the database is just an empty place holder) in particular causes the 
phylogenetic reconstruction to miss entirely the proximity of the Indo-Iranic and the 
Hellenic branches. Near the subtree shown in Figure~\ref{TochFig} one finds 
several instances of misplaced languages of the type discussed in \S \ref{proxSec} above. 

\begin{center}
\begin{figure}
\includegraphics[scale=0.75]{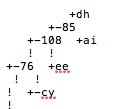}
\caption{Eastern Armenian:  cy = Eastern Armenian ($84\%$),
ee = Pima (misplaced Uto-Aztecan),
ai = Ossetic Digor, 
dh= Ossetic Iron.\label{ArmFig}}
\end{figure}
\end{center}

The situation with the Armenian branch is very problematic in the PHYLIP analysis
of the SSWL data. There are three entries recorded in the database: 
Western Armenian is $68\%$ mapped, while Eastern Armenian appears as
two different entries in the database, one $84\%$ mapped and the other only
$52\%$ mapped. Classical Armenian only appears as an empty place holder with 
no data in the current version of the database. These three data points are not
placed in proximity of one another in the PHYLIP reconstruction. Western Armenian
ends up completely misplaced (it appears in proximity of Korean and Japanese).
This misplacement may be corrected if one first subdivides data
by language families and then runs the phylogenetic reconstruction only on the
Indo-European data. The better mapped entry for Eastern Armenian is placed in proximity of  
the subtree of Figure~\ref{TochFig} containing the Tocharian--Anatolian
branch and some Indo-Iranian languages (plus some other misplaced languages
from other families). The nearest neighbors that appear in this region of the tree 
are Digor Ossetic and Iron Ossetic: again this is likely an effect of the poor mapping 
of the Indo-Iranic branch of the Indo-European family, as in the case of  
Figure~\ref{TochFig}. Another error due to misplacement from an entirely different
family occurs, with the Uto-Aztecan Pima placed in this same subtree, see Figure~\ref{ArmFig}.
This subtree is placed adjacent to a subtree containing a group of Balto-Slavic languages (and
some misplaced languages) with both of these branches then connecting to the
subtree of  Figure~\ref{TochFig}. The poorly mapped Eastern Armenian entry ($52\%$)
is placed as single leaf attached to an otherwise deep inner node of the tree. 
Another language that is often difficult to position in the Indo-European tree, Albanian ($68\%$ mapped), 
is misplaced in the PHYLIP reconstruction, and placed next to Gulf Arabic ($69\%$ mapped). 

\smallskip

These examples confirm the fact that a naive phylogenetic analysis of the SSWL database cannot
deliver any reliable information on the question of the early branchings of the Indo--European tree.

\medskip
\section{Phylogenetic Networks}

We verified that the same types of problems illustrated in the previous subsections occur when
the SSWL data are analyzed using phylogenetic networks instead of the PHYLIP phylogenetic trees.

\smallskip

We compiled the SSWL data \cite{SSWL}, using only the Indo-European languages, which have
more complete parameter information as a sample set. As in the tree analysis discussed before,
we input the syntactic parameters as a sequence of binary strings into the phylogenetic networks 
programs.

\smallskip

The {\tt Splitstree 4} program\footnote{{\tt http://ab.inf.uni-tuebingen.de/data/software/splitstree4/download/manual.pdf}} 
generated a split tree, which is intuitively a confidence
interval on trees. The farther from 'tree-like' the generated tree, the less 
any given tree is able to describe the evolution of the languages. The output of
this program indicated that the phylogenetics of languages analyzed on the basis
of SSWL syntactic parameters diverges strongly from being tree-like. As discussed
before, this may be regarded as further indication of systematic problems that
create high uncertainties in the candidate trees. These are again an illustration of
the effect of a combination of the factors (1)--(4) listed in \S \ref{ListSec}. 

\smallskip

We also fed the same data to the 
{\tt Network 5} program.\footnote{{\tt http://www.fluxus-engineering.com/Network5000\_user\_guide.pdf}}
This generated a phylogenetic network, using the median-joining
algorithm which represents all of the shortest-path length (maximum parsimony) trees
which are possible given the data. 

\smallskip

\begin{center}
\begin{figure}
\includegraphics[scale=0.45]{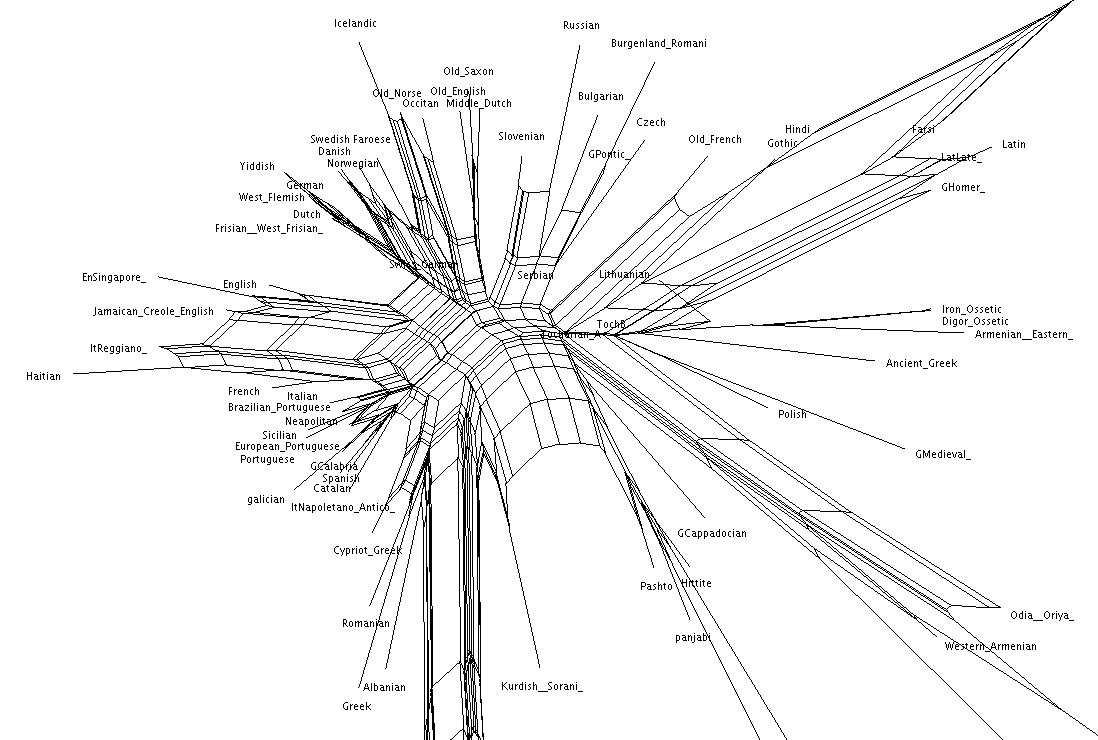}
\caption{Phylogenetic network produced by {\tt Splittree 4} on the entire SSWL database. \label{SSWLNetwork}}
\end{figure}
\end{center}

\begin{center}
\begin{figure}
\includegraphics[scale=0.5]{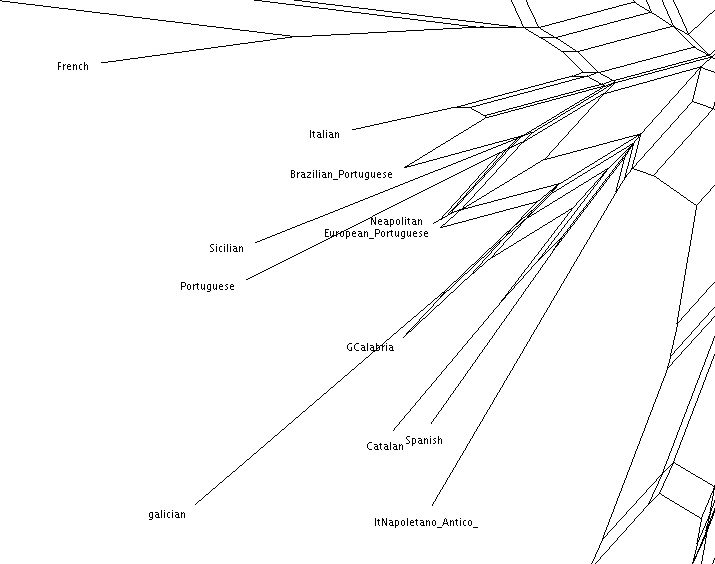}
\caption{Latin languages region of the phylogenetic network.\label{LatinNetwork}}
\end{figure}
\end{center}

\begin{center}
\begin{figure}
\includegraphics[scale=0.5]{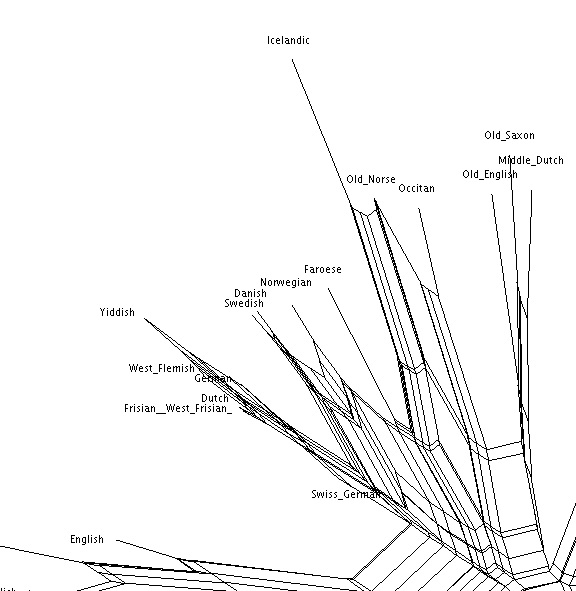}
\caption{Germanic languages region of the phylogenetic network.\label{GermanicNetwork}}
\end{figure}
\end{center}

We discuss below some of the aspects of the network generated by {\tt Splittree 4} in comparison
to some of the outputs described above obtained with the PHYLIP phylogenetic trees. 
Figure~\ref{SSWLNetwork} illustrates a large region of the phylogenetic network produced by
{\tt Splittree 4} using the entire set of SSWL data. It is evident that some of the same problems
we have discussed before occur in this case as well, in particular the misplacement of the
ancient languages with respect to their modern descendent (see the position of Latin and
Ancient Greek, for example). However, with respect to the PHYLIP results discussed above,
we see less instances of languages that get completely misplaced within the wrong family.
For example, as one can see from Figures~\ref{GermanicNetwork} and \ref{LatinNetwork},
Ancient Neapolitan now appears correctly placed in the Latin languages (and near 
Spanish) rather than misplaced among the Germanic languages as in Figure~\ref{GermFig}.
However, one can see that other problems that occurred in the PHYLIP reconstructions for
this group of languages are still present in the {\tt Splittree 4} network. For example, as in
Figure~\ref{LatinFig}, Portuguese appears closer to Italian than to Spanish in the network
of Figure~\ref{LatinNetwork}, contrary to the general understanding of the phylogenetic
tree of the Latin languages. (We will discuss the case of the subtree of the Latin languages
more in detail in \S \ref{AGtreesec} below.) Misplacements of languages within these
smaller subfamilies are still occurring, however: one can see that, for example, in 
the positioning of the Romance language Occitan in the region of the phylogenetic network
in proximity of Germanic languages like Old Norse and Icelandic in Figure~\ref{GermanicNetwork}.

\smallskip

The results of the {\tt Splittree 4} phylogenetic networks analysis of the Indo-European languages 
are available as the file {\tt Indo\_Euro.nex}, which can be downloaded at the URL \newline
{\tt http://www.its.caltech.edu/$\sim$matilde/PhylogeneticSSWL2}

\bigskip
\section{Phylogenetic Algebraic Geometry}\label{AGsec}

Given the unsatisfactory results one obtains in analyzing the SSWL database with
software aimed at phylogenetic reconstructions, one can turn the problem on its head
and try to obtain specific quantitative estimates of the level of reliability or unreliability
of specific subsets of the SSWL data for the purpose of phylogenetic, by relying on
existing reconstructions of linguistic phylogenetic trees, obtained by other linguistic
methods and other sources of data, which are considered reliable reconstructions.
The problem is then to test the distribution at the leaves of the tree obtained from
the SSWL data with specific polynomial invariants associated to a given reliable
tree. Such invariants would be vanishing on a probability distribution at the leaves
obtained from an evolutionary process modeled by a Markov model on the tree,
hence we can use the estimate of how far the values are from zero as a numerical
estimate of a degree of unreliability of the data for phylogenetic reconstruction.
Again, this does not identify explicitly the source of the problem, among the various
possible causes outlined above, but it still gives a numerical estimate that can
be useful in trying to improve the results. We propose here to use methods from
phylogenetic algebraic geometry to achieve this goal. We first give a quick
review of the main setting of phylogenetic algebraic geometry and then we
illustrate in some specific examples how we intend to use these techniques
for the purpose described here.

\subsection{Phylogenetic models: general assumptions}

The basic setup for linguistic phylogenetic models consists of a {\em dynamical process} 
of language change (which in our case means change of syntactic parameters), considered
as a Markov process on a {\em binary tree} (a finite tree with all internal vertices of valence $3$).

\smallskip

It can be argued whether trees really give the best account of language change 
based on syntactic data, rather than more general non-simply-connected graphs
(generally referred to as ``networks"). We will return to discuss some empirical 
reasons in favor of phylogenetic networks instead of trees in \S \ref{GeomSec} below. 
The mathematics of phylogenetic networks is discussed at length in \cite{Gus} and
\cite{Hus}. About the use of phylogenetic networks in Linguistics, see \cite{Nak}.

\smallskip

Another general assumption of phylogenetic models, which requires careful examination
in the case of applications to Linguistics, is the usual assumption that the variables 
(binary variables in the case of syntactic parameters) behave like  
{\em independent} identically distributed variables, whose dynamics evolves
according to {\em the same} Markov process. This assumption is especially
problematic when dealing with syntactic parameters because of the presence
of relations between parameters that are not entirely understood, so that it is
currently extremely hard to ensure one is using a set of independent binary
variables. Moreover, while acceptable in first approximation, even the
assumption that the underlying Markov model driving the change should be
the same for all syntactic parameters appears problematic. The fact that different
syntactic parameters have very different frequencies of occurrence among world
languages certainly suggests otherwise. We will return to this point in \S \ref{GeomSec}
and suggest a possible approach, based on the results of \cite{PBZMYKM}, to
correct, at least in part, for this problem. 

\smallskip

The leaves of the tree correspond to the modern languages with observed values of
the parameters giving a joint probability distribution
\begin{equation}\label{leavesP}
 \P(X_{\ell_1}=i_1, \ldots, X_{\ell_n}=i_n)= p_{i_1,\ldots, i_n}, 
\end{equation} 
with $i_k \in \{0,1\}$, and with $n$ the number of leaves. Here the
quantity $p_{i_1,\ldots, i_n}$ represents the frequency with which 
syntactic parameters of the languages $\ell_1, \ldots, \ell_n$ at the leaves 
of the tree have values $(i_1,\ldots,i_n)\in \{0,1\}^n$, respectively. 

\smallskip

In the usual setting of Markov models for phylogenetic reconstructions, one further assumes that
all the {\em inner nodes} are hidden variables and that only the distribution at the
leaves of the tree is known. Here again we encounter a problem with respect to
applications to Linguistics. In certain language families, like the Indo-European family,
several ancient languages have known parameters. In the SSWL database, for
instance, Ancient Greek is one of the very few languages that are 100$\%$ mapped
with respect to their list of 115 parameters. Thus, one needs to consider
some of the inner vertices as known rather than hidden. One way to do that is to 
consider a single leaf coming out of some of the inner vertices that will correspond to
the known values of the parameters at that vertex. As we discussed in \S \ref{TreeSec}
above, one encounters problems with the placement of the ancient languages in the
PHYLIP reconstruction of the syntactic phylogenetic trees, which should be corrected for.
Better results are obtained when one first separates out the data into language
families and subfamilies and builds trees for smaller subfamilies first, including the
known data about the ancient languages, and then combines these trees into a larger
tree. This procedure avoids the type of problem mentioned in \S \ref{TreeSec}, by which
the greater syntactic similarity between some of the ancient Indo-European languages
like Latin and Ancient Greek is detected correctly, but in turn prevents their respective placement
in the correct position with respect to the modern languages that originated from them. 

\smallskip

For a given set of $n$ leaves, there are 
$$ \tau_n = \frac{(2n-4)!}{(n-2)! 2^{n-2}} $$
different  possible binary tree topologies. Clearly, it
is not a computationally efficient strategy to analyze all of them.
However, one would like to have some computable invariants
that one can associate to a given candidate tree $T$, which estimates
how accurate $T$ is as a phylogenetic tree, among all the $\tau_n$
possible choices, given knowledge of the joint probability
distribution \eqref{leavesP} at the leaves. The Phylogenetic Algebraic
Geometry approach (see \cite{PaSturm}, \cite{PaSturm2} and the survey
\cite{Bocci}) aims at constructing such phylogenetic invariants using
Algebraic Geometry and Commutative Algebra. We review the
main ideas in the next subsection.

\smallskip
\subsection{Phylogenetic varieties and ideals}\label{AlgGeomSec}

We consider here the Jukes--Cantor model describing a Markov process on a
binary rooted tree $T$ with $n$ leaves. The stochastic behavior of the model
is determined by the datum of a probability distribution $(\pi, 1-\pi)$ at the
root vertex (the frequency of expression of the $0$ and $1$ values of the
syntactic parameters at the root) and the datum of a bistochastic matrix
$$ M^e=\begin{pmatrix} 1-p_e & p_e \\ p_e & 1-p_e \end{pmatrix} $$
along each edge of the tree. 
These data $(\pi, M^e)$ are often referred to in the literature as parameters of
the model. In order to avoid confusion with our use of the term parameter for the
syntactic binary variables, we will refer to the $(\pi, M^e)$ as ``stochastic 
parameters".  For a tree $T$ with $n$ leaves, and variables with $k$ states, the
number of stochastic parameters is
$$ N = (2n-3) k(k-1) + k-1. $$
In our case, with binary variables, we have $k=2$ and the number of stochastic
parameters of the model is simply $N=4n-5$. 

\smallskip

{\em Phylogenetic invariants} are polynomial functions $\phi$ that 
vanish on all the expected distributions $p_{i_n,\ldots, i_n}$
at the tails of the tree $T$, for all values of the stochastic parameters $(\pi, M^e)$.

\smallskip

The simplest example of such an invariant is the linear
polynomial
$$ \phi( z_{i_n,\ldots, i_n} ) = -1+ \sum_{i_n,\ldots, i_n} z_{i_n,\ldots, i_n}, $$
since the joint distribution at the leaves is normalized by 
$\sum_{i_n,\ldots, i_n} p_{i_n,\ldots, i_n} =1$. This invariant is uninteresting,
in the sense that it is independent of the tree $T$, hence it does not provide
any information about distinguishing between candidate phylogenetic trees.
In general one seeks other, more interesting, phylogenetic invariants $\phi_T$,
and the minimum number of such invariants required for phylogenetic inference.
An answer to this question is provided by Algebraic Geometry, as shown
in \cite{AllRho}, \cite{PaSturm}, \cite{PaSturm2}, \cite{SturmSull}. 

\smallskip

Consider the polynomial ring $\C[z_{i_1,\ldots,i_n}]$, where $n$ is the number of leaves of
the tree and and $i_k\in \{0,1\}$ for all $k=1,\ldots, n$. The 
phylogenetic invariants are defined by the vanishing $\phi_T(p_{i_1,\ldots,i_n})=0$.
This condition determines an ideal $\cI_T$ in the polynomial ring. For a Markov
model as above, with $N=4n-5$ stochastic parameters $(\pi, M^e)$, one obtains
a polynomial map
$$ \Phi: \C^{4n-5} \to \C^{2^n} $$
that assigns $\Phi(\pi,M^e)=p_{i_1,\ldots, i_n}$. This is, more explicitly, of the form
$$ p_{i_1,\ldots, i_n}= \Phi(\pi,M^e)=\sum_{w_v\in \{0,1\}} \pi_{w_{v_r}} \prod_e M^e_{w_{s(e)},w_{t(e)}}, $$
with a sum over ``histories" (paths in the tree) consistent with the data at the leaves. This determines
an algebraic variety, the {\em phylogenetic variety}, given by the Zariski closure
$$V_T=\overline{\Phi(\C^{4n-5})} \subset \C^{2^n}.$$ 
Dually we have a map $$\Psi: \C[z_{i_1,\ldots,i_n}]\to \C[x_1,\ldots,x_{4n-5}]$$ 
with ${\rm Ker}\Psi=\cI_T$, where $\cI_T$ is the {\em phylogenetic ideal}. 

\smallskip

One can use phylogenetic invariants to select between candidate 
phylogenetic trees in the following way. Suppose one obtains, through
some phylogenetic algorithm, a candidate phylogenetic tree $T$. One
also has available the joint probability distribution \eqref{leavesP} 
of the binary variables at the leaves. By evaluating phylogenetic
invariants $\phi_T \in \cI_T$ at the observed distribution $p_{i_n,\ldots, i_n}$,
one can check whether the candidate tree $T$ satisfies 
\begin{equation}\label{phiTest}
\left| \phi_T(p_{i_n,\ldots, i_n}) \right| < \epsilon 
\end{equation} 
for all phylogenetic invariants $\phi_T \in \cI_T$, and for a fixed error size $\epsilon$.
The candidate tree $T$ is an acceptable phylogenetic tree if and only if
the estimate  \eqref{phiTest} is satisfied. 
Geometrically, the test \eqref{phiTest} can be rephrased as the property that the point 
$p_{i_1,\ldots, i_n}\in \C^{2^n}$ is $\epsilon$-close to the phylogenetic variety 
$V_T$ if and only if $T$ is an acceptable phylogenetic tree. Computationally, this
method requires obtaining a set of explicit generators for the phylogenetic ideal $\cI_T$.

\smallskip

In the case of the Jukes--Cantor model with $k=2$, it was proved in \cite{SturmSull}
that the phylogenetic ideal $\cI_T$ is generated by polynomials of degree two. A
completely explicit set of generators for the Jukes--Cantor model with $k=2$ was obtained
in \cite{AllRho}, where it is proved that phylogenetic ideal $\cI_T$ generated by the $3\times 3$-minors
of all {\em edge flattenings} of the tensor $P=(p_{i_1,\ldots,i_n})$. The edge flattenings are
defined by the following procedure. Start with a tree $T$ with Markov model $(\pi, M^e)$ and 
with $P\in \C^{2^n}$ the joint probability distribution $P=(p_{i_1,\ldots, i_n})$ at the $n$ leaves.
The choice of an edge $e$ in a tree $T$ with $n$ leaves determines two
connected components of $T\smallsetminus \{ e \}$, hence two sets of leaves
$\{ \ell_1,\ldots, \ell_r\}$ and $\{ \ell_{r+1}, \ldots, \ell_n \}$. Thus, the $2^n$ binary variables
at the $n$ leaves are partitioned into a set of $2^r$ variables and a set of $2^{n-r}$ variables,
and the joint distribution $P=(p_{i_1,\ldots, i_n})$ determines a $2^r\times 2^{n-r}$-matrix $Flat_{e,T}(P)$
specified by setting 
$$ Flat_{e,T}(P)(u,v) = P(u_1,\ldots, u_r, v_1, \ldots, v_{n-r}). $$
It can be shown that the rank of this matrix is ${\rm rank}( Flat_{e,T}(P)) \leq 2$ (for binary variables, $k=2$), 
hence all $3\times 3$ minors of the matrix must vanish. It is shown in \cite{AllRho} that, for $k=2$ any 
number $n$ of leaves, the phylogenetic ideal $\cI_T$
is generated by the $3\times 3$ minors of the matrices $Flat_{e,T}(P)$ of all edge flattenings. It is easy
to see that, even for small trees, there is a very large number of these $3\times 3$ minors, hence
the number of generators of the phylogenetic ideal grows rapidly with the size of the tree.

\smallskip

Note that, while for the purpose of validating a candidate phylogenetic tree $T$ it would
be necessary to check that all these generators of the phylogenetic ideal vanish (or
nearly vanish as in \eqref{leavesP}), in order to invalidate a candidate tree it sufficed to
find at least one of these $3\times 3$ minors for one of the flattenings that evaluates 
on the observed joint distribution $P=(p_{i_1,\ldots, i_n})$ to a value larger than the
allowed error size $\epsilon$.

\section{Phylogenetic invariants and syntactic trees} \label{AGtreesec}

In this section we show how phylogenetic invariants can be used to improve the
phylogenetic tree reconstructions based on SSWL syntactic data. 

\subsection{Phylogenetic invariants of small syntactic trees}

We focus here on sufficiently small subtrees of the syntactic phylogenetic tree of languages
compiled from the SSWL data, for which the computation of phylogenetic invariants becomes
feasible. Using phylogenetic invariants, we compare the small trees obtained in this way with
phylogenetic trees obtained by other linguistic methods and considered reliable, so as to
estimate the validity of the joint distribution at the leaves obtained from SSWL data. 

\smallskip

We present here an example, based on the subtree of the Latin languages within the
Indo--European family. A more detailed analysis of other subtrees of the Indo--European
family will be presented elsewhere. 

\smallskip

We have seen in \S \ref{wrong1sec} that the naive PHYLIP analysis of the entire SSWL
database misplaces Portuguese in the subtree of the Indo-European languages that
collects the Latin languages. We have also seen in \S \ref{wrong4sec} that the same
analysis misplaces Latin, separating it from the tree of the modern languages
that originated from it. 

\smallskip

We now perform a more accurate analysis, still using only the SSWL data, but where
we use the {\em a priori} knowledge of the grouping of certain languages into a 
subfamily. Thus, we select only the languages {\em Latin, Italian, French, Spanish, Portuguese}.

\smallskip

\begin{center}
\begin{figure}
\includegraphics[scale=0.5]{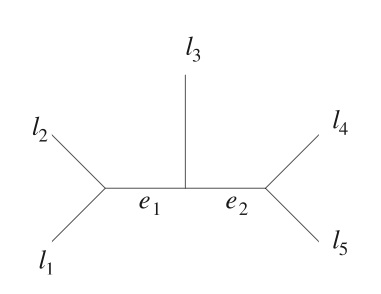}
\caption{Tree topology for the phylogenetic tree of the Latin languages, with $\ell_1=$ French,
$\ell_2=$ Italian, $\ell_3 =$ Latin, $\ell_4=$ Spanish, $\ell_5=$ Portuguese.\label{treeFlatP}}
\end{figure}
\end{center}

The phylogenetic tree that is generally agreed, through other linguistic reconstructions, to
best represent the relative position of these languages would be a tree topology as
illustrated in Figure~\ref{treeFlatP}. Note that this is also the tree reconstruction for this
group of languages obtained in \cite{LongGua} using a set of syntactic parameters different
from those recorded in the SSWL database.

The tree of Figure~\ref{treeFlatP} has 
two possible splits: $\{ \ell_1, \ell_2 \}\cup \{ \ell_3, \ell_4, \ell_5\}$ and
$\{ \ell_1, \ell_2, \ell_3 \}\cup \{ \ell_4, \ell_5 \}$. The corresponding flattenings
are given by the matrices 
\begin{center}
\includegraphics[scale=0.5]{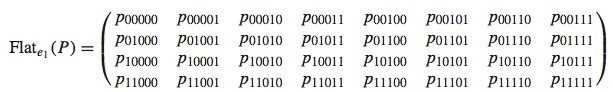}
\includegraphics[scale=0.5]{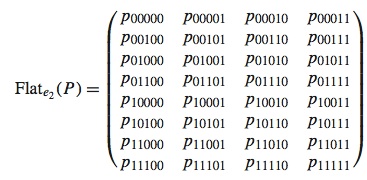}
\end{center}
where the $p_{i_1,i_2,i_3,i_4,i_5}$ are the frequencies of the observed binary variables
at the ends, under the assumption that these behave like independent equally
distributed random variables, evolving according to the same Markov model on the tree.

\smallskip

Using the data of SSWL parameters for these five languages reported in the Appendix,
we obtain matrices ${\rm Flat}_{e_1}(P)$ and ${\rm Flat}_{e_2}(P)$ of the form
$$ {\rm Flat}_{e_1}(P) = \begin{pmatrix}
\frac{31}{106} & \frac{1}{106} & \frac{1}{106} & 0 & \frac{23}{106}  & \frac{3}{106} & 0 & \frac{1}{53} \\
\frac{1}{106} & 0 & 0 & \frac{1}{106} & 0 & \frac{1}{106} & 0 & \frac{3}{106} \\
\frac{5}{106} & 0 & \frac{1}{53} & 0 & 0 & 0 & 0 & 0 \\
\frac{1}{53} & 0  & \frac{1}{106} & \frac{4}{53} & 0 & 0 & 0 & \frac{21}{106} 
\end{pmatrix} $$
$$ {\rm Flat}_{e_2}(P) = \begin{pmatrix}
\frac{31}{106} & \frac{1}{106} & \frac{1}{106} & 0 & 0 \\
\frac{23}{106} & \frac{3}{106} & 0 & \frac{1}{53} & 0 \\
\frac{1}{106} & 0 & 0 & \frac{1}{106} & 0 \\
0 & \frac{1}{106} & 0 & \frac{3}{106} & 0  \\
\frac{5}{106} & 0 & \frac{1}{53} & 0 & 0 \\
0 & 0 &0 &0&0 \\
\frac{1}{53} & 0 & \frac{1}{106} & \frac{4}{53} & 0 \\
0 & 0 & 0 & \frac{21}{106} & 0
\end{pmatrix}. $$
Evaluating all the $3\times 3$ minors of these matrices with {\tt Maple} and selecting
the maximum absolute value of the resulting phylogenetic invariants gives  
\begin{equation}\label{maxphylinv}
\max \left| \phi_T (p_{i_1,\ldots,i_5}) \right| = \frac{2415}{1191016} = 0.0020277.
\end{equation}

\smallskip

The fact that for the tree of Figure~\ref{treeFlatP} the distribution at the leaves determined by the
SSWL parameters is extremely close to being a zero of all the phylogenetic invariants implies
that the SSWL parameters are in fact in very good agreement with the recognized correct
topology of the phylogenetic tree, but only when the set of languages is previously restricted to
a smaller subfamily and {\em only the SSWL parameters that are fully mapped for that
subfamily are taken into account}. 

\smallskip

This result seems to indicate that the main source of the problems we encounter when doing
a naive phylogenetic analysis using the entire SSWL database are not necessarily due to an
intrinsic problem with the SSWL data (that is, it is not primarily due to problem number (3) in
the list in \S \ref{ListSec}). It seems rather that the problems encountered above stem from a
combination of problems (1), (2), and (4). The use of the phylogenetic invariants method
bypasses problem (1), while the prior restriction to a smaller subfamily bypasses problems (2)
and (4). A more detailed analysis of this approach with phylogenetic invariants, applied to
other language subfamilies using SSWL data will be carried out more extensively elsewhere.

\medskip
\section{Dependencies and Geometry}\label{GeomSec}

As we already mentioned above, the problem of the construction
of reliable syntactic phylogenetic trees is closely related to the problem of
relations and dependencies between syntactic parameters. 
Are there universal relations that hold across all languages?
Are there relations that depend on language families? Can these relations
be expressed geometrically, as is the case with relations between
continuous coordinates that give rise to topological or differentiable manifolds?
Are there different geometries associated to different language families?
How detectable are relations between syntactic parameters computationally?
Recently, a mathematical approach to these questions was proposed
in \cite{Mar}, \cite{PBZMYKM}, \cite{PGGCLDM}, \cite{STM}. 

\smallskip

In \cite{PGGCLDM}, it was shown, again using SSWL data, that
syntactic parameters of different language families have different
persistent homology. The persistent generators of $H_0$ appear to
correspond to a subdivisions of a given language family into
major subfamilies, such as, for example, the Indo-Iranic and the European 
subfamilies of the Indo-European family, or the Mande, Atlantic-Congo, and
Kordofanian subfamilies of the Niger-Congo family. A persistent generator
of the $H_1$ was found in the case of the Indo-European family. It appears
to be related to the position of the Hellenic branch in the Indo-European family.
It is presently unclear whether this reflects the effect of a genuine historical-linguistic
phenomenon, such as an influence of Ancient Greek, at the syntactic level,
upon some other European languages (such as some of the Slavic languages),
whether it detects the presence of homeoplasy in syntactic parameters,
or whether it is due to the nature and format of the syntactic data collected in
the SSWL database. However, the presence of non-trivial persistent
generators of the $H_1$ in the persistent homology of the data set 
is a strong indicator that networks (non-simply-connected graphs) and not
trees may provide a better topology for syntactic phylogenetic linguistics. 

\smallskip

In \cite{PBZMYKM}, it was shown that, to some extent, the presence
of dependencies between the syntactic parameters listed in the SSWL
database can be detected using Kanerva networks. The latter were
introduced in \cite{Kanerva} as sparse distributed memories aimed at
modeling associative memory in neuroscience. It is well known that,
in fact, Kanerva networks are very useful for reconstructing corrupted data
and detecting the degree of recoverability of certain parts of the data as
a function of the remaining ones. In particular, this makes them suitable
for detecting the presence of relations between data. It was shown in
\cite{PBZMYKM} that different syntactic parameters in the SSWL
database exhibit different degrees of recoverability in a Kanerva
network. An overall effect can be identified, which depends on the
frequency with which a certain syntactic parameter is expressed
across world languages. This effect can be reproduced using
random data with the same frequencies. However, there is an
additional effect that can be detected normalizing with respect to
the frequency and that should be a genuine expression of the
level of dependence of a particular syntactic parameter upon
the remaining ones. The resulting normalized score computed
in \cite{PBZMYKM} is therefore a numerical estimate of the
degree of dependence/independence of a given binary
syntactic variable. The presence of these computationally
detectable dependence relations affects some of the fundamental
assumptions of the Markov models of phylogenetic trees, in
particular the assumption that all the binary variables are independent,
identically distributed variables. A possible way to compensate for
this problem in the model it to consider a weighted version of the
joint probability distribution $P=p_{i_1.\ldots, i_n}$ at the leaves
of the phylogenetic tree, where the frequency of expression of
the parameters is computed in such a way that each parameter is
weighted according to  the corresponding normalized degree of 
recoverability in a Kanerva network, in such a way that the
independent variables are weighted more than the dependent
ones. This restores the fact that the independent variables 
assumption of the Markov model can be at least approximately
satisfied. 

\bigskip

\section*{Appendix: the SSWL parameters of the Latin languages}

The phylogenetic invariants for the tree of Latin languages of Figure~\ref{treeFlatP}
are evaluated at the probability distribution $p_{i_1,i_2.i_3,i_4,i_5}$ at the leaves,
based on the SSWL parameters for this group of languages. There are 106 parameters
in the SSWL database that are completely mapped for all of these five languages.
We have excluded from the list all those SSWL parameters that are only mapped for some
but not all of the languages in this group. With the notation $\ell_1=$ French,
$\ell_2=$ Italian, $\ell_3 =$ Latin, $\ell_4=$ Spanish, and $\ell_5=$ Portuguese,
the syntactic parameters are given by the following list. The column on the left
lists the SSWL parameters $P$ as labeled in the database, \cite{SSWL}.

\begin{center}
\includegraphics[scale=0.5]{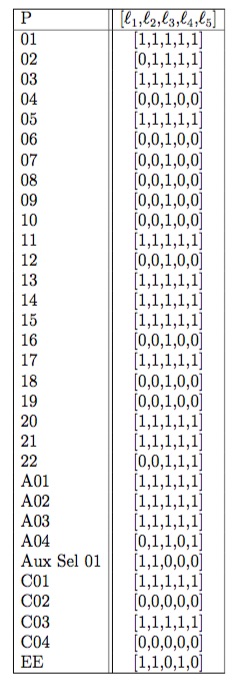}
\includegraphics[scale=0.5]{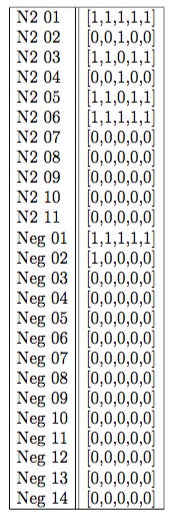}
\includegraphics[scale=0.5]{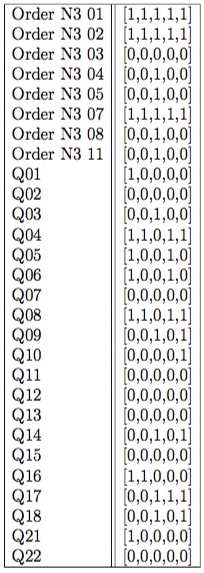}
\includegraphics[scale=0.5]{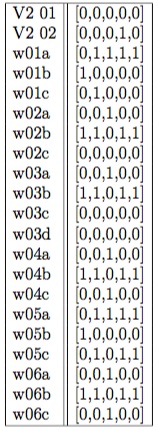}
\end{center}

One can see by inspecting the different groups of parameters in this list
that several parameters within the ``same group" tend to behave in the same
way (e.g.~all the {\em Neg} parameters) or in more highly correlated way
than across groups of parameters. This observation is consistent with the
more general observation of dependencies observed through the Kanerva
networks method in \cite{PBZMYKM}. Thus, in order to better fit this set of binary
variables with the hypothesis of independent equally distributed variables in
Markov processes, it may be better to select a subset of the SSWL parameters
that cuts across the various groups of more closely correlated variables. We will
discuss this aspect more in details elsewhere. 

\smallskip

The probability $p_{i_1,i_2.i_3,i_4,i_5}$ is then computed by counting the frequencies of
occurrence of binary vectors $[i_1,i_2,i_3,i_4,i_5] \in \{0,1\}^5$ among the 106 vectors of SSWL 
parameters above. The only nonzero frequencies are
$$ p_{0,0,0,0,0}=\frac{31}{106}, \ \ \ p_{0,0,0,0,1}=\frac{1}{106}, \ \ \  p_{0,0,0,1,0}=\frac{1}{106}, \ \ \ p_{0,0,1,0,0}=\frac{23}{106}, $$
$$ p_{0,0,1,0,1}= \frac{3}{106}, \ \ \ p_{0,0,1,1,1}=\frac{2}{106}, \ \ \  p_{0,1,0,0,0}=\frac{1}{106}, \ \ \ p_{0,1,0,1,1}=\frac{1}{106}, $$
$$ p_{0,1,1,0,1}= \frac{1}{106}, \ \ \  p_{0,1,1,1,1}=\frac{3}{106}, \ \ \ p_{1,0,0,0,0} = \frac{5}{106}, \ \ \ p_{1,0,0,1,0}=\frac{2}{106}, $$
$$ p_{1,1,0,1,0}=\frac{1}{106}, \ \ \ p_{1,1,0,0,0}=\frac{2}{106}, \ \ \ p_{1,1,0,1,1}=\frac{8}{106}, \ \ \ 
p_{1,1,1,1,1}=\frac{21}{106}. $$

Note how these frequencies confirm some well known facts about the Latin languages.
Syntactic parameters (as recorded in SSWL) are very likely to have remained the
same across all five languages in the family, with a higher probability of a feature not
allowed in Latin remaining not allowed in the other languages ($31/106$) than of a
feature allowed in Latin remaining allowed in the other languages ($21/106$). It is also
very likely that a feature is the same in all the modern ones but different from Latin, with
a much higher incidence of cases of a feature allowed in Latin becoming disallowed
in all the other languages ($23/106$) than the other way around ($8/106$). Among the
remaining possibilities, we see incidences where French has an allowed feature
that is missing in the other languages ($5/106$) of disallowed ($3/106$) and cases
where Latin and Portuguese have the same feature allowed, which is disallowed in
the other languages ($3/106$): all other nonzero entries have only two or less occurrences. 
The resulting matrices for the edge flattenings of the tree of Figure~\ref{treeFlatP} are then
as computed in \S \ref{AGtreesec}.

\end{document}